\pdfoutput=1
\documentclass[11pt]{article}

\usepackage[preprint]{acl}

\usepackage{times}
\usepackage{latexsym}

\usepackage[T1]{fontenc}
\usepackage{xcolor}
\usepackage{hyperref}
\usepackage{graphicx}
\usepackage{amsmath}
\usepackage{subcaption}
\usepackage{makecell} % For line breaks in cell text
\usepackage{url}
\usepackage{booktabs}
\usepackage{amssymb}
\usepackage{adjustbox}
\usepackage{wrapfig}
\usepackage[utf8]{inputenc}

\usepackage{microtype}
\usepackage{inconsolata}

\usepackage{booktabs}
\usepackage{adjustbox}
\usepackage{colortbl}
\usepackage{pgf}

\usepackage{xfp}

\usepackage{inconsolata}

\usepackage{textcomp}
\usepackage{graphicx}

\newcommand{\ifeval}{\textsc{IFEval}}
\newcommand{\dolly}{\textsc{Dolly}}
\newcommand{\evol}{\textsc{Evol}}
\newcommand{\flan}{\textsc{FLAN}}
\newcommand{\alpaca}{\textsc{Alpaca}}

\newcommand{\llmbar}{\textsc{LLMBAR}}
\newcommand{\alpacaeval}{\textsc{AlpacaEval}}
\newcommand{\openllm}{\textsc{OpenLLM}}

\newcommand{\selectedmodels}{\ensuremath{M_{\text{selected}}}}
\newcommand{\randombaselines}{\ensuremath{M_{\text{random}}}}
\newcommand{\fulldatamodels}{\ensuremath{M_{\text{full-dataset}}}}

\newcommand{\strictrandom}{\ensuremath{S_{\text{strictrandom}}}}
\newcommand{\Sstrategy}{\ensuremath{S_{\text{strategy}}}}
\newcommand{\cherry}{\ensuremath{S_{\text{cherry}}}}
\newcommand{\longest}{\ensuremath{S_{\text{longest}}}}
\newcommand{\deita}{\ensuremath{S_{\text{deita}}}}
\newcommand{\alpacasus}{\ensuremath{S_{\text{alpagasus}}}}

\newcommand{\random}{\ensuremath{S_{\text{random}}}}

\newcommand{\redstar}{{\color{red}$\star$}}

\usepackage[most]{tcolorbox} % Include the tcolorbox package

\definecolor{lightbrown}{rgb}{0.91, 0.84, 0.72}

\newtcolorbox{takeaway}{
  colback=lightbrown, % Background color of the box
  colframe=black, % Frame color
  boxrule=0.2mm, % Frame thickness
  arc=2mm, % The arc of the box corners
  top=2mm, % Top margin within the box
  bottom=2mm, % Bottom margin within the box
  left=2mm, % Left margin within the box
  right=2mm, % Right margin within the box
  boxsep=0mm, % Space between text and frame in all directions
}

\title{Chasing Random -- Investigating the  \textit{\textgravedbl}Gains\textit{\textacutedbl} achieved through Instruction Selection Strategies at Scale}
\author{Harshita Diddee$^{1}$ \quad Daphne Ippolito$^{1,2}$ \\ $^{1}$Carnegie Mellon University \\ $^{2}$ Google Deepmind\\ {\tt \small \{hdiddee,dippolit\}@andrew.cmu.edu}}

\begin{document}
\maketitle
\begin{abstract}
Prior work \cite{Zhou2023LIMALI} has shown that language models can be tuned to follow user instructions using only a small set of high-quality instructions. This has accelerated the development of methods that filter a large, noisy instruction-tuning datasets down to high-quality subset which works just as well. However, typically, the performance of these methods is not demonstrated across a uniform experimental setup and \textit{thus their generalization capabilities are not well established}.
In this work, we analyze popular selection strategies across different source datasets, selection budgets and evaluation benchmarks:
Our results indicate that selection strategies generalize poorly, often failing to consistently outperform even random baselines. We also analyze the cost-performance trade-offs of using data selection.
Our findings reveal that data selection can often exceed the cost of fine-tuning on the full dataset, yielding only marginal—and sometimes no gains compared  to tuning on the full dataset or a random subset.

\end{abstract}

\section{Introduction}
% \ssd{finetuning -> fine-tuning}
% \ssd{preparing -> training}preparing
Instruction fine-tuning is often considered a crucial step in training large language models, LLMs, to effectively meet the needs of users. By training LLMs over tens of thousands instruction-response tuples that highlight user preferences, models can demonstrate \textit{instruction-following capabilities} which position them as useful tools for a wide variety of tasks. There has been a rapid increase in the development of \textit{instruction selection} strategies \cite{qin2024unleashingpowerdatatsunami, wang2024surveydataselectionllm} to curate a subset of high-quality instructions to train competitive instruction following models more efficiently.  

% Typically, these strategies select a relatively small number of "high quality" instruction-response tuples from large instruction tuning datasets such that a model trained on this subset demonstrates instruction following capabilities comparable to another model trained on the entire dataset. 

\begin{figure}[t!]
    \includegraphics[width=\linewidth]{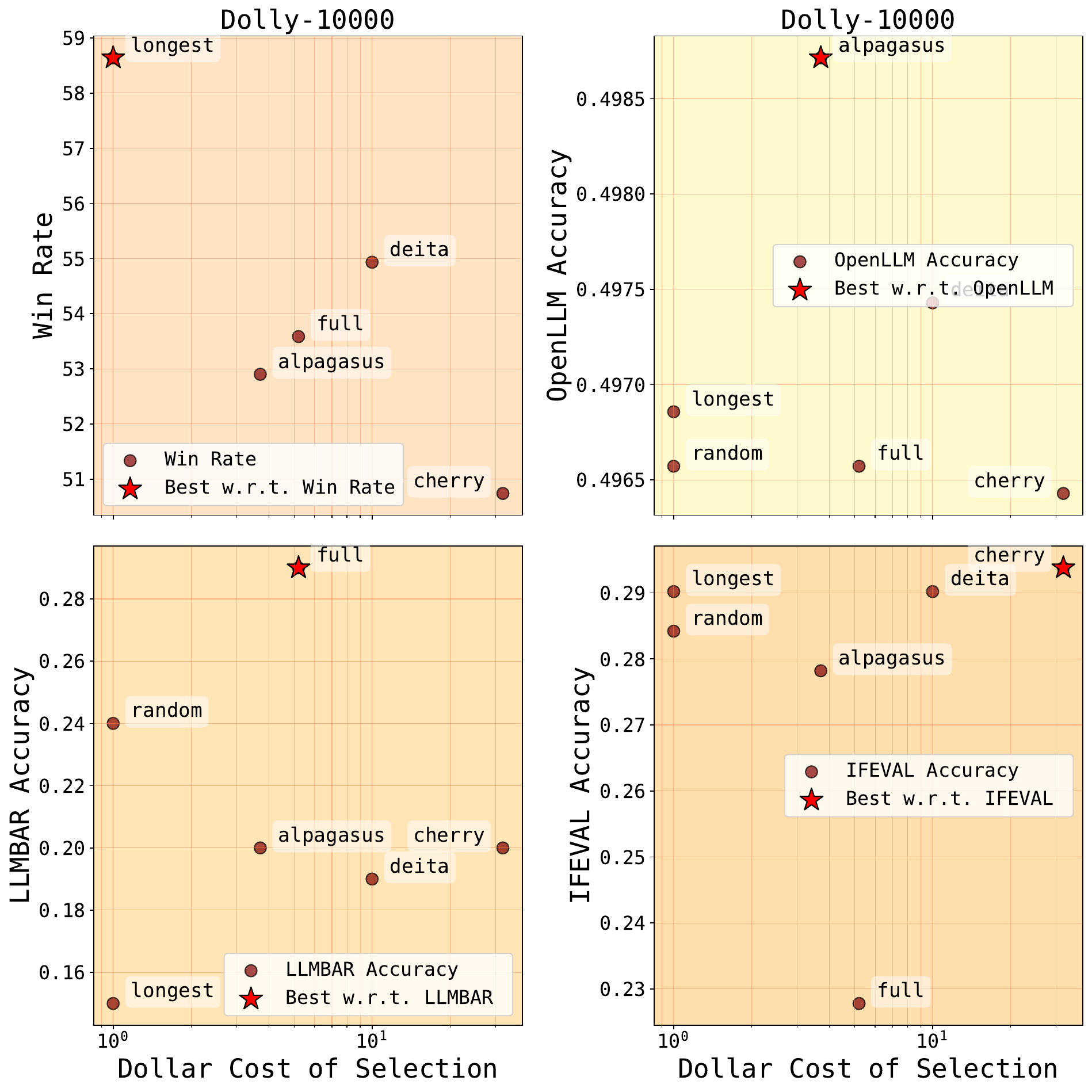}
    \caption{Selection Cost Versus Performance on different benchmarks when selecting 10000 samples from \dolly~\cite{DatabricksBlog2023DollyV2}: Upper Left Region (low cost, high performance) is ideal. Key Takeaways are: (a) Random baselines are reasonably competitive whilst incurring the least cost (b) Depending on the evaluation metric, the best strategy varies significantly with the setup (\redstar indicates best selection strategy on the benchmark).}
    \label{fig:teaser}
\end{figure}

% \href{https://huggingface.co/datasets/databricks/databricks-dolly-15k}{DOLLY}
% \ssd{cite instead of href}
% \ssd{increase font size, plot looks blurred (try dpi=300 if you haven't already), use [t!] for positioning}

The experimental setups of general-purpose instruction data selection can be very varied \cite{qin2024unleashingpowerdatatsunami}; Unlike task-specific data selection, they are not geared towards optimizing performance for some specific goal \cite{xie2023data}. Therefore, their utility is strongly tied to their generalization beyond a few limited setups endorsed by their designers. 

% \yz{I don't understand the above sentence. Also diverse has a positive connotation, and you may want to highlight these setups are diverse in a bad way ---non-uniform maybe?}
% Neither are they significantly limited by some intractable scale as in the case of pretraining data selection \cite{albalak2024surveydataselectionlanguage},
Measuring this generalization is hard for several reasons. Firstly, proposed strategies are often applied across arbitrary experimental setups including different source datasets, selection budgets and testing benchmarks. Additionally, since there isn't a single set of behaviors expected of instruction tuned models, it is unclear performance gained through selection on one instruction-following benchmark will induce correlated gains on other instruction-following benchmarks. Finally, the time and resources necessitated by a selection strategy vary significantly. While strategies like \citet{chen2023alpagasus} can directly incur dollar-cost through their dependence on API-accessible commercial large language model APIs, others like \cite{Li2023FromQT, Liu2023WhatMG} design selection methods that involve finetuning or inferencing on LLMs, thus mandating a GPU-reliant infrastructural cost.
% others like \cite{Li2023FromQT, Liu2023WhatMG} can mandate an infrastructural cost to perform GPU-reliant subsampling on the entire instruction tuning dataset.
% \yz{A reader may wonder where this GPU cost comes from. Could be more direct and say what the costs are in such methods?}

\textbf{Contribution:} In this work, we carry out an exhaustive investigation for over 60 experimental configurations across 4 evaluation benchmarks each to provide evidence for the following findings (a) Instruction Selection Strategies don't generalize to reasonably similar experimental configurations. Consequently, \textbf{no selection strategy beats random selection \textit{consistently}}. (b) \textbf{Competence in General Purpose Instruction Following is a subjective goal} and hence, comparing selection strategies on different facets of this goal can produce contradictory trends. (c) \textbf{Many strategies scale poorly as the budget of selection increases}: Incurred selection costs can often overshoot the cost of training with the entire dataset and do not give consistently high gains over random selection carried out at negligible cost. 

We argue that the lack of generality and consistent performance over a naive baseline makes it difficult to use existing example selection strategies in the wild: if selection through a strategy is not \textit{consistently cost-effective} over a naive form of subsampling (random-sampling) across \textit{reasonably similar experimental configurations}, it is unclear if selection is an advantageous step in the process of training competitive LLMs.

% all \yz{did you actually try all possible configurations? might want to tone down here; or just say you tried $N=?$ combinations}

 % \item \textbf{Instruction Selection Strategies fail to generalize to custom experimental configurations}: The correlation between a selection strategy's performance between any 2 experimental configurations across any given benchmark is poor, indicating that gains accrued by a strategy on a setup are not applicable to even reasonably similar setups. We posit that this failure of generalization is exacerbated by the inherent differences between the instruction-following benchmarks which are used to measure and compare the performances of sampling strategies. 
\section{Literature Review}
% \yz{Is there a reason why this section isn't called related work? I don't have a problem with it but just curious lol}
% \deicomment{Add a paragraph here giving a bit of background on instruction selection datasets, and citing Table 1}
 % \ssd{the prev sent is off} 
We focus on general-purpose instruction selection methods, which aim to equip models to follow user queries that aren't specific to a fixed task, capability or domain \cite{wang2024surveydataselectionllm}.Such methods involve strategies including rule-based metrics \cite{Cao2023InstructionMW}, length \cite{Zhao2024LongIM}, diversity \cite{Liu2023WhatMG} and model derived uncertainity measurements \cite{Li2023FromQT} to subsample large instruction tuning datasets. They sit in contrast to task-specific data selection strategies which optimize for performance on a known test distribution or task specification \cite{Xia2024LESSSI, Xie2023DataSF, pan-etal-2024-g}. 

Due to broad definition of instructing following, experimental setups for such work can show significant variation: While some adopt source distiributions of varying origins \cite{Zhou2023LIMALI, li2024superfilteringweaktostrongdatafiltering, shen2024rethinkingdataselectionsupervised}, some even synthetically augment subsets of data during their selection process \cite{Liu2023WhatMG}. Similarly, the selection budget applied on these datasets can vary from anywhere between a mere 200 samples \cite{wei2023instructiongpt4200instructionparadigmfinetuning} to 15K \cite{Du2023MoDSMD, Xia2024LESSSI}. We summarize some of the most popular choices in Table \ref{tab:source-evaluation-description} and utilize these for our experiments. \\
A few methods also \cite{mekala-etal-2024-smaller, Xia2024LESSSI} acknowledge and accordingly attempt to address the relatively high cost of selection by either exploring the use of cheaper proxies like smaller models, low-rank approximations or adopting sequential pipelines \cite{ge2024clusteringrankingdiversitypreservedinstruction} to make instruction data selection more efficient. 
Finally, work like \cite{Liu2024TakeTE, wang2024surveydataselectionllm} have highlighted the concerns in comparing between the performance of instruction selection strategies. Through a unified comparison based on efficiency and feasibility, \citet{Liu2024TakeTE} provide strong evidence that the comparison between instruction selection strategies needs to be more holistic. Distinctive from this work though, their evaluation does not focus on a comparison including random baselines. % \ssd{you are highlighting existing work, but try to write couple of lines like this. Our work is different ... So that reader has less cognitive load}
\begin{takeaway}
\textit{\textbf{Our Work}} Distinct from prior work, we focus on calibrating both, the performance gain and the cost benefit of various instruction selection strategies against the negligible cost alternative, random baselines. In the process, we also uncover the sharp sensitivity of selection strategies to their experimental setups which can significantly harm the ease of their adoption. 
\end{takeaway}

\begin{table*}[ht]
\centering
\small
\begin{tabular}{lccp{0.35\linewidth}}
\toprule
\textbf{\begin{tabular}[c]{@{}l@{}}Source\\ Distribution\end{tabular}} & \textbf{Authorship} & \textbf{\begin{tabular}[c]{@{}c@{}}Number of \\ Samples\end{tabular}} & \multicolumn{1}{c}{\textbf{Brief Description}} \\
\midrule
\flan{} ~\cite{Longpre2023TheFC} & Automatic & 88k & Includes Flan 2021, P3, Super-Natural Instructions among other datasets. \\
\dolly{} ~\cite{DatabricksBlog2023DollyV2} & Human & 15k & Instruction-responses crafted by Databricks employees. \\
\evol{} ~\cite{Xu2023WizardLMEL} & LLM & 196k & Modifying seed instructions using ChatGPT. \\
\alpaca{}~\cite{alpaca} & LLM & 52k & ChatGPT\footnote{\href{https://platform.openai.com/docs/models/moderation}{text-davinci-003}} driven generation with Self-Instruct's pipeline. \\
\midrule
\textbf{\begin{tabular}[c]{@{}l@{}}Evaluation \\ Setup\end{tabular}} & \textbf{\begin{tabular}[c]{@{}c@{}}Number of \\ Samples\end{tabular}} & \textbf{Metric} & \textbf{Brief Description} \\
\midrule
\ifeval{}~\cite{Zhou2023InstructionFollowingEF} & 500 & \begin{tabular}[c]{@{}c@{}}Instruct, Prompt\\ Level Accuracy\end{tabular} & Instructions have verifiable prompts to check if model fulfills all prompts in an instruction. \\
\alpacaeval{} ~\cite{alpaca_eval} & 805 & \begin{tabular}[c]{@{}c@{}}Length Controlled\\ Win Rate\end{tabular} & Judges LLM responses by an automatic annotator with high human-correlation. \\
\llmbar{} ~\cite{Zeng2023EvaluatingLL} & \begin{tabular}[c]{@{}c@{}}100\\ (Natural Set)\end{tabular} & Accuracy & Checks model preference over instruction responses to check if a model identifies faithful responses. \\
\openllm{} ~\cite{eval-harness} & Task-Specific & Accuracy & MMLU, ARC-Easy, ARC-Challenge, WinoGrande, TruthfulQA, HellaSwag \\
\bottomrule
\end{tabular}
\caption{A brief overview of the source distributions we investigate and the Evaluation Setups we consider.}
\label{tab:source-evaluation-description}
\end{table*}

\section{Experimental Setup}
\label{dataset-desription}
In this section, we briefly describe our source datasets, the selection strategies we study and our evaluation setup.  

\subsection{Source Datasets and Evaluation Setups} 
Table \ref{tab:source-evaluation-description} provides a concise description of all our datasets and evaluation benchmarks. For FLAN, as a precaution against including disproportionately high representations towards tasks that are overly-represented in the original composition, we curate a smaller subset of FLAN, by limiting the datapoints sampled per task to 50 for our evaluation. The resulting dataset contains 88K examples and we refer to this version as \flan{}. For \alpacaeval{}, we use a fixed randomly sampled subset of 300 samples to reduce the cost overhead of our evaluations. We use the default recommended annotator configuration using GPT-4-Turbo. 

\subsection{Selection Strategies}
% \ssd{use noindent or paragraph uniformly}
% \ssd{consider using "noindent textbf\{\}" to save space}
\paragraph{Alpagasus (\alpacasus{})} \citet{chen2023alpagasus} use GPT-3.5 as scorer (between 1-5) to score samples from \alpaca{} and include the highest scoring samples. 
\paragraph{Longest (\longest{})} \citet{Zhao2024LongIM} include the instructions with the longest responses. 
\paragraph{Cherry (\cherry{})} \citet{Li2023FromQT} use a sequential approach of selecting instructions: they apply k-means clustering to the last hidden state embeddings of all instruction in a source dataset to get a set of 1000 instructions (100 clusters and 10 samples per cluster). Then, they use this subset of instructions to finetune a model, referred to as the pre-experienced model. Finally, this model scores each sample with an Instruction Difficulty or IFD and the highest scoring samples are included in the selected subset. 
\paragraph{DEITA (\deita{})}\citet{Liu2023WhatMG} train a scorer akin to Alpagasus to first score the entire dataset cheaply and then, rather than choosing the entire budget of instructions in one shot - iteratively construct the selected subset by checking the similarity of a candidate instruction to the current pool of instructions.
\paragraph{Uniform Random (\random{})} This is the naivest form of sampling and acts as our baseline. We report numbers with error bars for trials across 3 such random seeds. We also resample for any random subset that ends up having more than 30\% overlap with the data sampled with any strategy for all datasets expect dolly (due to Dolly's limited size, a maximum overlap of about 50\% is possible only for the highest budget 10000).
\paragraph{Strict Random (\strictrandom{})} We also create a special variant of our random-baselines called the "strictrandom" baselines which is created by sampling from the dataset after removing \textbf{all} the target instructions that have been deemed high-quality by \textit{any} of the selection strategies. In practice, the strictrandom baselines can also be considered as sampling data from the complement set of all strategies' "high-quality" subsets of budget 10000. 

% \yz{This is a very cool baseline --- maybe call this `` random among discarded''?}
\paragraph{Full Dataset:}  The entire dataset is used to train the model. Note that we include this variant without tuning optimally for each dataset and include this only to compare the gains that can be naively procured by avoiding selection altogether.
% \ssd{add version and cite}
\subsection{Base Model}
We use the LLaMa-7B \cite{Touvron2023LLaMAOA} for all our experiments.This model's choice is dictated by it's use as a common choice for demonstrating and ablating the performance of the instruction selection strategies that we study (Table 2 \cite{qin2024unleashingpowerdatatsunami}). We provide all details of the 3 hyperparameter sets we test in the Appendix~\ref{hyperparameters}). 

\section{Results}
In this section, we present evidence supporting our conclusions on the brittle generalization of instruction selection strategies (\S\ref{sec:random-baselines}  and \S\ref{sec:instruction-following-performance}) as well as the negative utility of expending cost on data selection \S\ref{sec:cost-of-selection}. 
% \yz{I think the intro here should be very short and punchy. One sentence suffices: say first nothing generalizes, and then hidden costs of example selection do not justify reduced training time.}
% In this section, we conduct experiments that demonstrate that the gains achieved through instruction data selection may be less impactful as previously deemed because of the following reasons: In \S\ref{sec:random-baselines} we discuss how random baselines in this space may be underestimated by experimental design, clouding their competence with respect to the studied strategies;  Following this we study how changes in the evaluation setup can impact our assessment of a selection strategy's performance \S\ref{sec:instruction-following-performance}. Finally, in \S\ref{sec:cost-of-selection}, we show how the cost incurred in selection can often be comparable to the cost of tuning on the entire data without a consistent guarantee on performance. Throughout the section, we use \selectedmodels, \randombaselines{} and \fulldatamodels{} to connote models trained on strategy-guided, randomly subsampled and the entire data respectively. 

\subsection{Most Strategies Fail to Beat Random Sampling Consistently}
\label{sec:random-baselines}
\begin{figure}[t!]
    \begin{subfigure}[b]{\linewidth}
        \includegraphics[width=\linewidth]{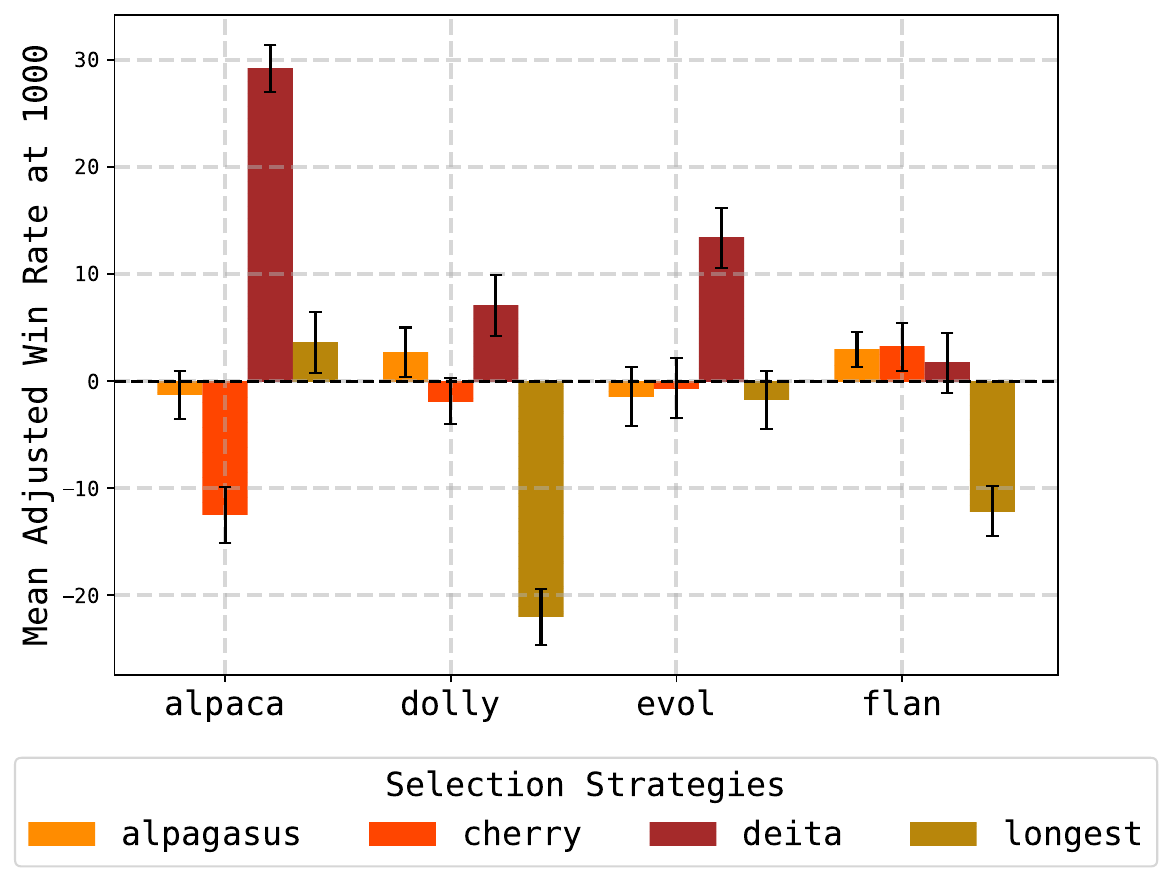}
        \caption{}
        \label{fig:alpacaeval-1k}
    \end{subfigure}
    \begin{subfigure}[b]{\linewidth}
        \includegraphics[width=\linewidth]{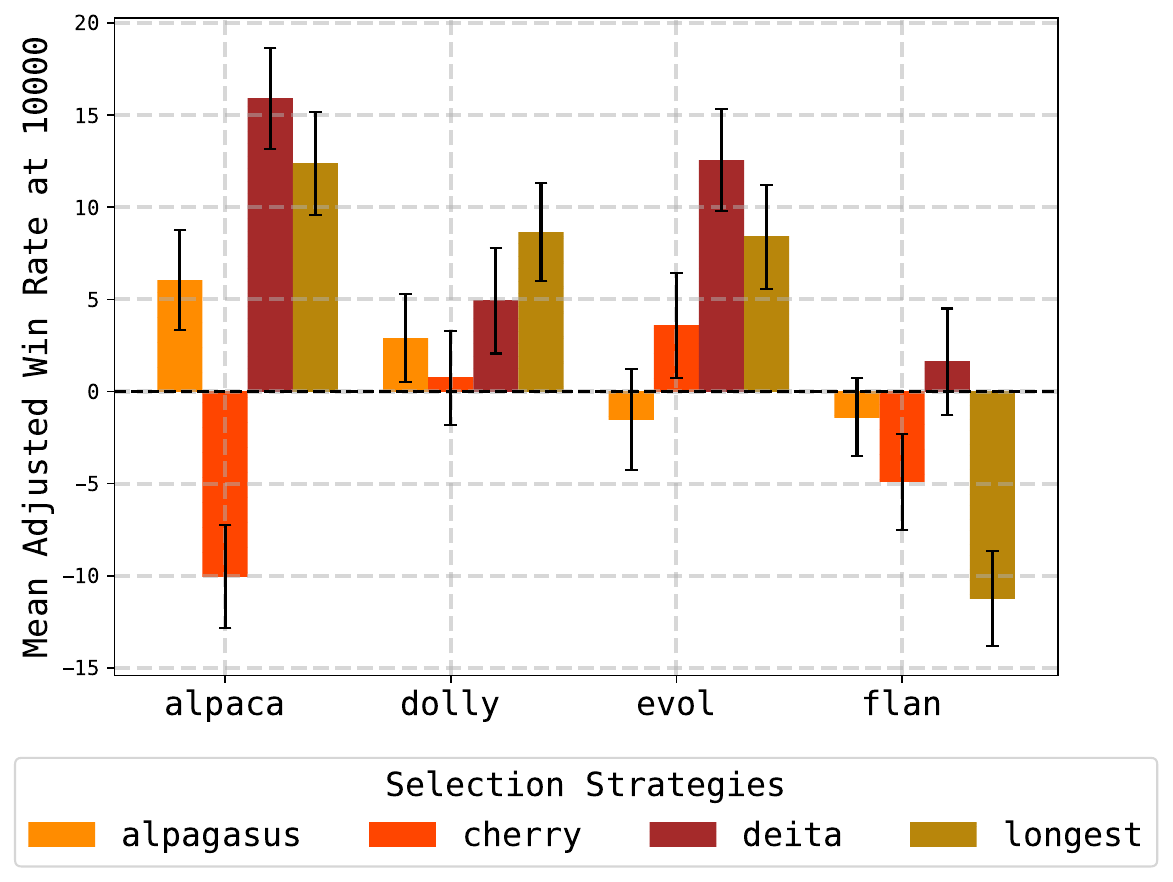}
        \caption{}
        \label{fig:alpacaeval-10k}
    \end{subfigure}
    \caption{Mean Adjusted Win Rates on \alpacaeval{} for budgets (a) 1000 (b) 10000. A bar along the negative y-axis indicates that the \randombaselines{} responses are preferred more than 50\% of the time by GPT-4. No strategy except \deita{} beats random baselines consistently. No strategy shows consistent performance trends across budgets as well (Section \S\ref{sec:random-baselines}) for more details.}
    \label{fig:alpacaeval-combined}
\end{figure}

% \ssd{use \{\} at the end of the variable for rendering the space in the pdf}
 In the space of instruction data selection, it is very common to show that \selectedmodels{} outperform \fulldatamodels{} by over 50\% (i.e., an LLM judge prefers the outputs of the \selectedmodels{} more than the \fulldatamodels{}). We modify this experimental setup to perform these comparisons between the \randombaselines{} and \selectedmodels{} on \alpacaeval{}. Specifically, for each model in \selectedmodels{}, we pair the output of the \selectedmodels{} with a randomly chosen inference generated by a random baseline from the \randombaselines{} trained for the same budget and dataset. We then compute the Mean-Adjusted Win-Rate by taking the signed difference between the win-rate\footnote{In all our experiments we use length controlled win-rate to negate the effects of length-bias in LLM judges.} of the \selectedmodels{} from 50\%. Our results across two budgets are summarized in Figure \ref{fig:alpacaeval-combined}. \\

% Example: Consider \longest trained \selectedmodels{} thoroughly underperform (loses by over 20\% from \randombaselines{} at a low budget of 1000 but dominate over the random baseline (by over 10\%) when scaled to 10000.

% \deicomment{This looks a lot better. My only comment would be to increase all the font sizes.}

% \ssd{See detailed results in Sec.~\ref{XX}. Consider adding this to add figs and tables where relevant.}

 % AlpacaEval is a popular instruction following benchmark that uses Win-Rate to discern if the quality of responses from a model under examination are better than some reference outputs

\noindent \textbf{Findings on \alpacaeval{}} No strategy except \deita{}, consistently dominates over the \randombaselines{} across all experimental configurations. To illustrate the practical implications of this observation, consider an NLP practitioner who intends to apply data selection on the \dolly{} dataset with a budget of 10,000 samples. They evaluate the performance of various selection strategies on \dolly{} at a smaller budget of 5,000 samples and conclude that \cherry{} is the most effective strategy (Figure \ref{fig:alpacaeval-combined}). However, when this strategy is applied and empirically tested at the intended budget of 10,000 samples, the results are the opposite: \cherry{} delivers the lowest performance among all strategies (Figure \ref{fig:alpacaeval-combined}). While we give an example with \cherry, it is reasonable to assume that other strategies experience similar inflection points in their performance with the change in budget. For example, even though \deita consistently outperforms random in this evaluation, it loses nearly 15\% of its dominance over \randombaselines{} at budget 10000 (when scaled from 5000) indicating the potential for an inflection point in performance for some larger budget.

\begin{takeaway}
\textit{\textbf{Takeaway}} This evaluation exemplifies that the performance estimate for a selection strategy is heavily influenced by the budget and source datasets on which it is tested, and purported gains may not transfer consistently across selection budgets or data sources.
\end{takeaway}

\begin{figure}[t!]
\centering
    \begin{subfigure}[b]{0.9\linewidth}
    \includegraphics[width=0.9\linewidth]{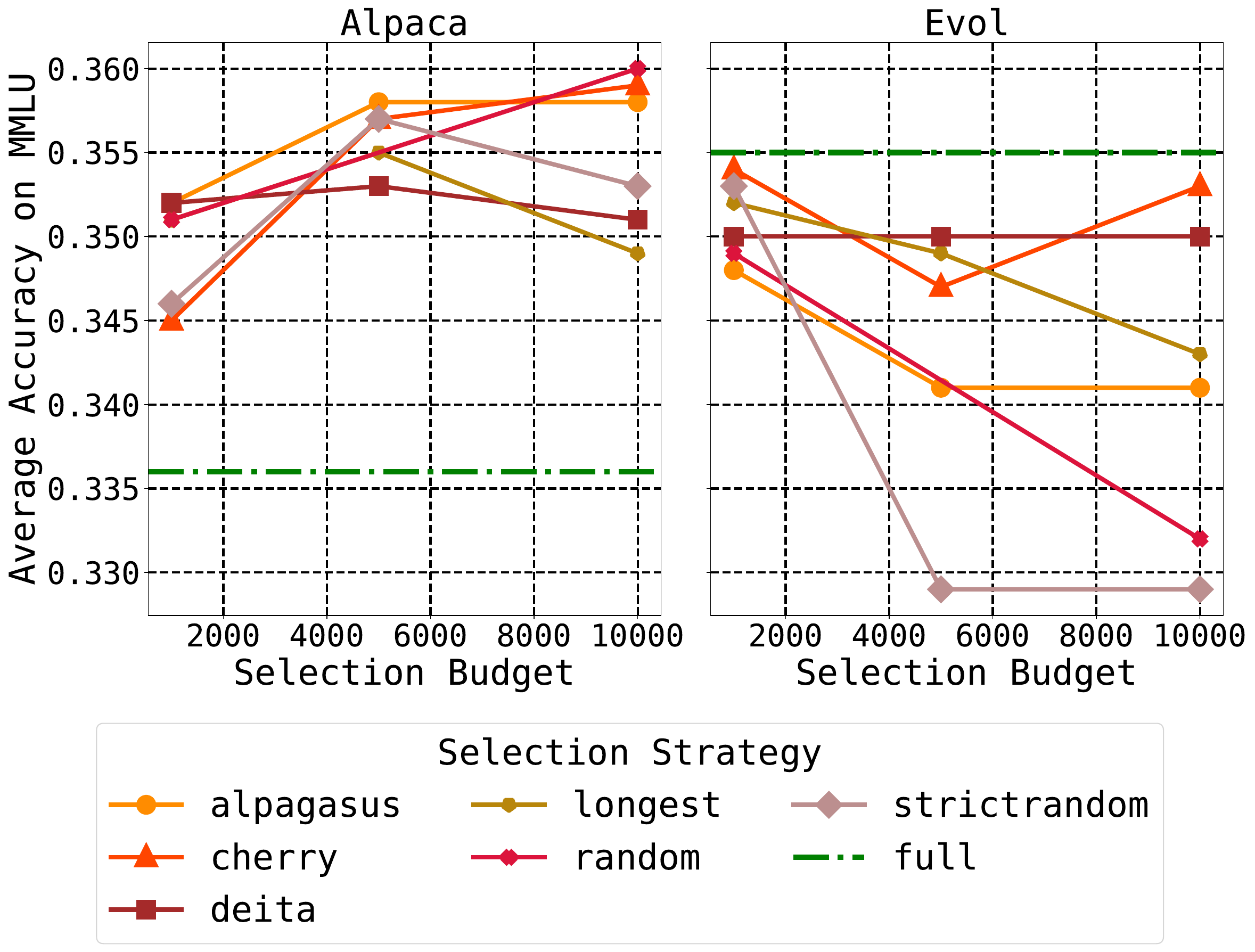}
    \caption{Performance Trends According to MMLU}  
        \label{fig:openllm}
    \end{subfigure}
    \hfill
    \begin{subfigure}[b]{0.9\linewidth}
    \includegraphics[width=0.9\linewidth]{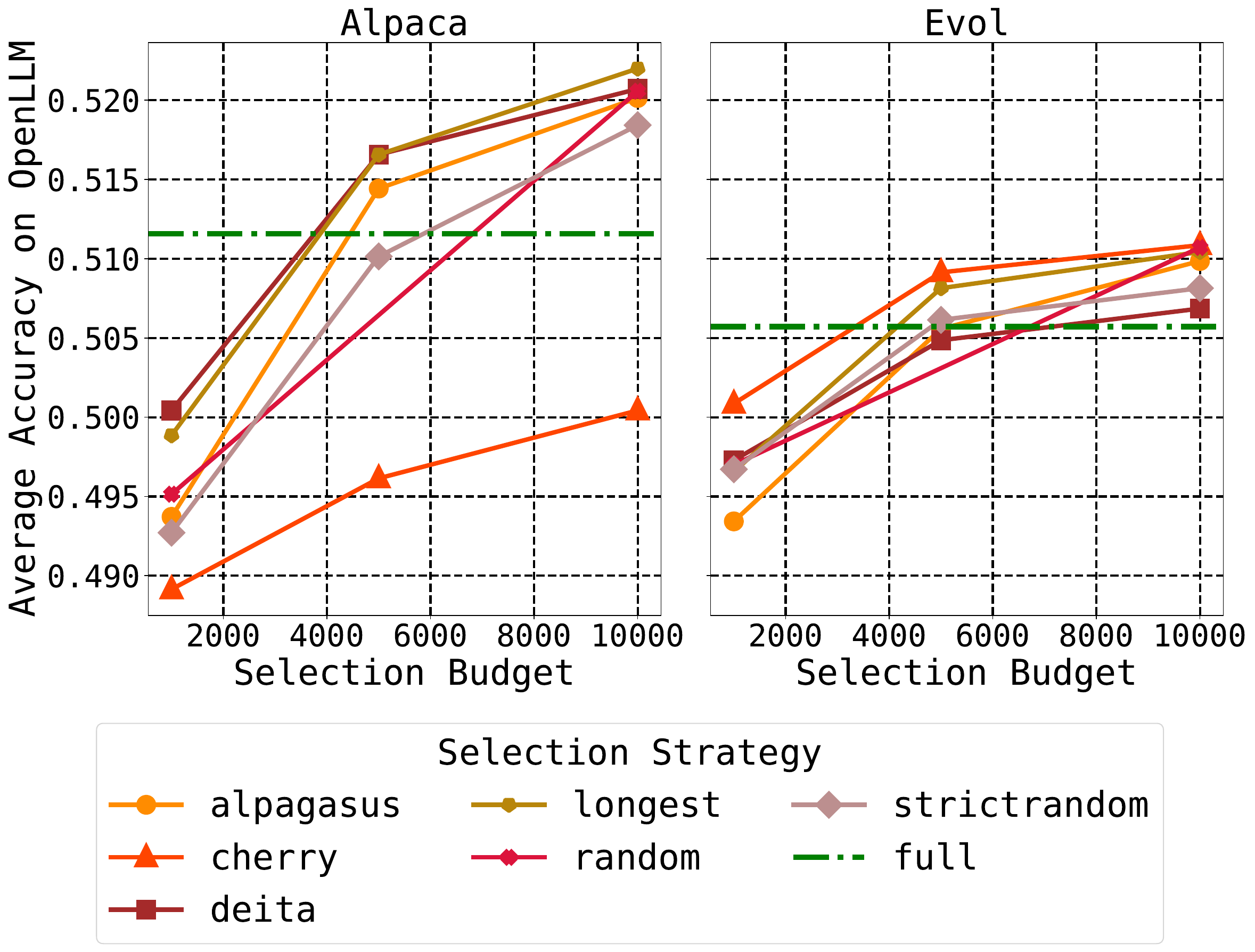}
    \caption{Performance Trends According to Average Performance on Subset of OpenLLM}
        \label{fig:openllm}
    \end{subfigure}
    \caption{There is a stark difference between the performance trends of selection strategies depending upon what subset of \openllm{} tasks are chosen for evaluation. \random{} is the worst performing strategy across all datasets when performance is gauged on MMLU, while \random{} shows competitive performance as more tasks from \openllm{} are considered. Details in \S\ref{openllm-results} and \ref{fig:openllm-all}.
    \label{fig:openllm}}.
\end{figure}

\begin{figure*}[t!]
\centering
\includegraphics[width=0.9\linewidth]{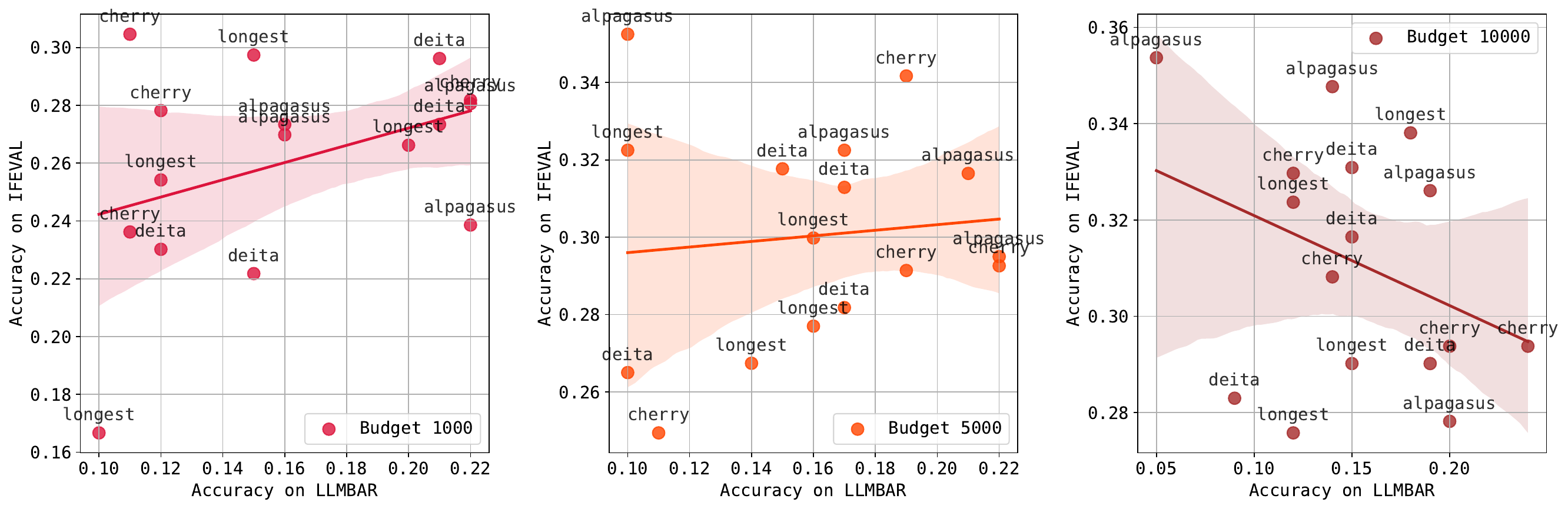}
        \caption{Mean Instruct-Level Accuracy of \selectedmodels{} on \ifeval{} versus Win-Rate on \alpacaeval{}:  The correlation between Win-Rate and \ifeval{} is entirely non-existent or weakly correlated at best. As budget increases these also appear to diverge: as performance drops on Win-Rate as \ifeval{} accuracy improves. (\S\ref{sec:instruction-following-performance} for further details.)}
        \label{fig:iffeval}
\end{figure*}

\paragraph{Findings with \openllm{}}
\label{openllm-results} To corroborate this trend, we evaluate \selectedmodels{} with \randombaselines{} on \openllm{}.  In Figure  \ref{fig:openllm}, we demonstrate the performance of \selectedmodels{} across different budgets on both (a) MMLU (the only task evaluated by \cite{chen2023alpagasus}) and (b) average performance on 7 tasks from OpenLLM (the largest union of tasks considered by \cite{Zhao2024LongIM, Li2023FromQT}).
Not surprisingly, we find extreme divergence in the observed performance trends of selection strategies depending upon which setup is adopted: While \random{} subsampling performs the worst by a significant margin against all selection strategies when evaluated using only MMLU (Fig \ref{fig:openllm} (a)), it performs far more competitively when more tasks from OpenLLM are considered (Fig \ref{fig:openllm} (b)), especially performing competitively at larger budgets. Note that this setup only highlights the difficulty arising out of using a non-standard subset of evaluation tasks and does not question if its even appropriate to consider \textit{any} of these tasks as a reasonable indicator of a model's instruction following capabilities. MMLU, for example, has been shown to demonstrate several contextual limitations \cite{Gema2024AreWD} in addition to being a multiple choice format task which significantly deviates from the traditional long-form generation format of instruction following benchmarks. Hence, it would not be too unreasonable to assume that it may not be a sufficiently aligned choice for demonstrating that a \selectedmodels{} demonstrates instruction following capabilities in the first place.  

% dels{} demonstrate instruction following capabilities in the first place. 
% Empirically, the performance trend of strict-random shows this too: actively excluding samples that are deemed high quality from selection can definitely reduce performance of even \randombaselines{}.

\begin{figure}[t!]
\small
\includegraphics[width=0.9\linewidth]{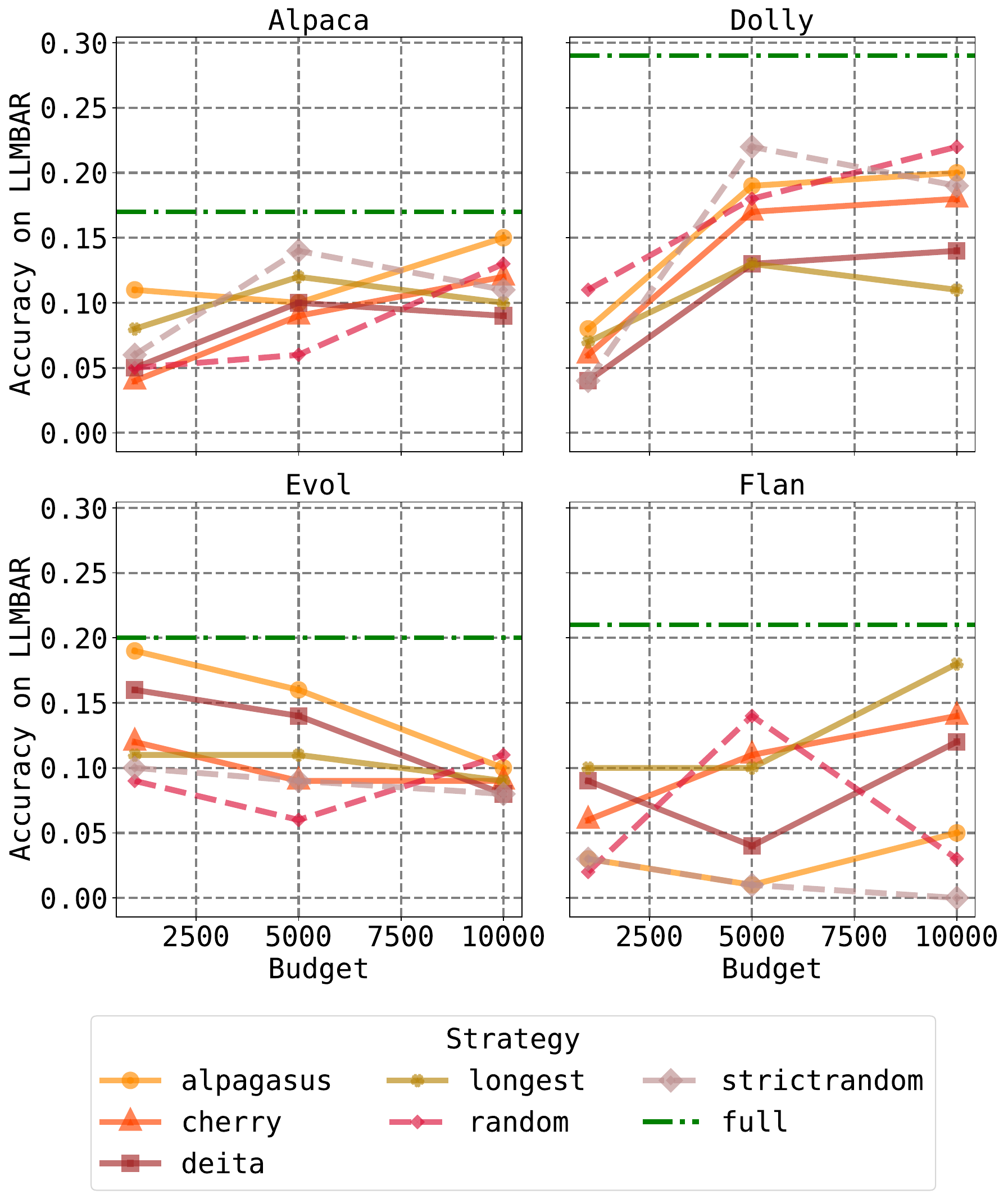}
    \caption{Performance on \llmbar{}: Both \randombaselines{} and \selectedmodels{} consistently underperform \fulldatamodels{}. }
    \label{fig:llmbar}
\end{figure}

% \yz{I think the takeaway for OpenLLM experiments fall under this section. Maybe merge?}
% \yz{I don't understand this section header. Consider rewrite without symbols?}

\subsection{Measuring Instruction Following for \selectedmodels produces contradictory trends}
\label{sec:instruction-following-performance}

Measuring instruction following capabilities is generally more complex than task-specific accuracy evaluation as instruction following models are expected to demonstrate a wide range of capabilities \cite{lou2024largelanguagemodelinstruction}. Consequently, the subjectivity in the coverage of topics and the performance ranges of each instruction following benchmark can further influence our estimates of a selection strategy's performance. Recently, an emerging class of benchmarks recommend evaluating models with instructions which have more objective requirements \cite{qin2024infobenchevaluatinginstructionfollowing, Zhou2023InstructionFollowingEF}. Accordingly, we conduct an evaluation of \selectedmodels{} on another popular instruction following benchmark that complies with this format, \ifeval{} \cite{Zhou2023InstructionFollowingEF}. \ifeval{} defines its own metrics, prompt-level and instruction-level accuracy, to measure how well a model response covers all the requirements delineated by each prompts and ultimately the test instruction. As in our previous evaluation with AlpacaEval and OpenLLM, we compare the performance of \selectedmodels{} and \randombaselines{} on this benchmark. 

 % \ssd{ctrl+f replace IFFEval with \ifeval{}}
 % which is designed to measure the instruction following capabilities at this granular level: each test instruction in this benchmark contains multiple prompts, fulfilling each of which is necessary to generate the complete and correct response for that instruction. 

% \yz{This is a really strong result! You should bold the cheapeast overall methods.}

% On 3 out of 4 datasets, no \selectedmodels{} consistently beats \randombaselines{} and even on the fourth, \randombaselines{} is the 2nd best strategy amongst all strategies.  

\paragraph{Findings from \ifeval{}} We include complete results on \ifeval{} in the Appendix (Figure \ref{fig:iffeval-all}), where we observe similarly competitive performance from \randombaselines{}; Here, we highlight another interesting observation derived through this benchmark: In Figure \ref{fig:iffeval}, we show the correlation between the Win-Rates for \selectedmodels{} and their \ifeval{} accuracy \ref{fig:iffeval}. The performance trends on both benchmarks appear very weakly correlated for our lowest budget, and show almost negative correlation after scaling \selectedmodels{} to the larget budget. This is particularly concerning as both benchmarks are widely used as indicators of instruction following capabilities and hence, at least by definition it is hard to pick the conclusions of one over the other. The practical implication of this correlation is observed when these setups disagree on what the most appropriate selection strategy for a setup is. In this case for instance, we see that a lower budget (Figure \ref{fig:iffeval}(, we can claim with reasonable confidence that \deita{} would give high performance across both benchmarks but as budget scales the trade-off between the performances on both benchmarks increases significantly making it hard to conclude which strategy has higher utility. We also observe similarly poor correlations between the performance trends of selection strategies when we do a pair wise comparison between other studied benchmarks like \openllm{} and \llmbar{} (Figure \ref{fig:correlation-btw-benchmarks}).

% For example, while Win-Rate has generally shows positive correlation with human judgements, recent research by \cite{Zheng2024CheatingAL} demonstrates that even a "null model" with constant output can achieve top-ranked win rates. This suggests one potential reason for using human judgment as a sole validation metric. 

% Multifaceted nature of instruction following: Different evaluation strategies may emphasize or enhance performance across various aspects of instruction following. A single benchmark or collection of benchmarks may not comprehensively capture all these facets.

To demonstrate this more concretely, we conduct a final evaluation on another instruction following benchmark, \llmbar{}.
\paragraph{Findings from \llmbar{}} In Figure \ref{fig:llmbar}, we observe that both \selectedmodels{} and \randombaselines{} perform poorly on this benchmark. Interestingly, unlike all other benchmarks we study where \fulldatamodels{} are either comparable in performance or even underperform \selectedmodels{}, on \llmbar{} we clearly see consistent performance improvement when the model is trained on the entire data. This result, hence, sits in complete contrast to all other benchmark evaluations as it exposes another facet of evaluation where selection is not advantageous at all. 

\begin{takeaway}
    \textit{\textbf{Takeaways}} Benchmarks do not show agreement on the performance trends of selection strategies (\S\ref{sec:iffeval}). Further, choosing representative tasks that are aligned with a subjective measure of instruction following can significantly alter the observed performance trends (seen through \S\ref{sec:openllm}). In such a case, it seems more useful to focus on data selection when we have prior objectives to optimize for as in test-distribution or task-specific selection. 
\end{takeaway}

% \hdtodo{[[UPDATE PLOTS WITH THE FULL FINETUNING PERFORMANCE PLOTS]]}

% Evaluating general purpose instruction data selection is subjective due to its dependence on an inherently subjective goal; Accordingly, it might make more sense to focus on data selection where we have prior objectives to optimize for (test-distribution or task-specific selection).

% We abstain from commenting on the inherent superiority of any evaluation strategy or benchmark, especially through means of correlating with human judgements because evidence shows that such assessments may also be fallible: For instance, while Win-Rate is shown to have positive correlation with human judgements - \cite{Zheng2024CheatingAL} show that even a "null model" that always outputs a constant response can achieve top-ranked win rates. Additionally, it is also possible that different strategies favor or boost performance of \selectedmodels{} across the multiple facets of instruction following, some of which may not be covered by an arbitrary collection of benchmarks. This benchmark judges the competence of an instruction tuned model by measuring its ability to discern which of the two supplied responses is a faithful response for a test instruction.

\subsection{Cost of Instruction Data Selection is Non-Trivial when compared to the cost of Tuning on the Entire Data}
\label{sec:cost-of-selection}

\setlength{\tabcolsep}{4pt} 
\begin{table*}[]
\small 
\centering
\begin{adjustbox}{width=\textwidth}
\begin{tabular}{cccclccccc}
\toprule
\rowcolor[HTML]{D2B48C}
\textbf{Dataset} &
  \textbf{\begin{tabular}[c]{@{}c@{}}Samples \\ (as multiples \\ of 1k)\end{tabular}} &
  \textbf{Alpagasus} &
  \multicolumn{3}{c}{\cellcolor[HTML]{D2B48C}\textbf{Cherry}} &
  \multicolumn{2}{c}{\cellcolor[HTML]{D2B48C}\textbf{DEITA}} &
  \multicolumn{2}{c}{\cellcolor[HTML]{D2B48C}\textbf{\begin{tabular}[c]{@{}c@{}}Entire\\ Dataset\end{tabular}}} \\ 
  \midrule
\textbf{\begin{tabular}[c]{@{}c@{}}Costing \\ Categories\end{tabular}} &
   &
  {\color[HTML]{343434} \begin{tabular}[c]{@{}c@{}}API Inference \\ Cost (USD)\end{tabular}} &
  \multicolumn{2}{c}{{\color[HTML]{343434} \begin{tabular}[c]{@{}c@{}}Rent Time\\ (min)\end{tabular}}} &
  {\color[HTML]{343434} \begin{tabular}[c]{@{}c@{}}Rent Cost \\ (USD)\end{tabular}} &
  {\color[HTML]{343434} \begin{tabular}[c]{@{}c@{}}Rent Time \\ (min)\end{tabular}} &
  {\color[HTML]{343434} \begin{tabular}[c]{@{}c@{}}Rent Cost\\ (USD)\end{tabular}} &
  \multicolumn{1}{c}{{\color[HTML]{343434} \begin{tabular}[c]{@{}c@{}}Rent Time\\ (min)\end{tabular}}} &
  \multicolumn{1}{c}{{\color[HTML]{343434} \begin{tabular}[c]{@{}c@{}}Rent Cost\\ (USD)\end{tabular}}} \\ 
EVOL &
  200 &
  \cellcolor[HTML]{D2B48C}50 &
  \multicolumn{2}{c}{3290} &
  \cellcolor[HTML]{D2B48C}427 &
  1000 &
  \cellcolor[HTML]{D2B48C}130 &
  1620 &
  \cellcolor[HTML]{D2B48C}216 \\
ALPACA &
  52 &
  \cellcolor[HTML]{D2B48C}12.66 &
  \multicolumn{2}{c}{855} &
  \cellcolor[HTML]{D2B48C}{\color[HTML]{000000} 111.15} &
  260 &
  \cellcolor[HTML]{D2B48C}33.8 &
  120 &
  \cellcolor[HTML]{D2B48C}15.6 \\
DOLLY &
  15 &
  \cellcolor[HTML]{D2B48C}3.7 &
  \multicolumn{2}{c}{246.75} &
  \cellcolor[HTML]{D2B48C}32.07 &
  75 &
  \cellcolor[HTML]{D2B48C}9.975 &
  40 &
  \cellcolor[HTML]{D2B48C}5.2 \\
FLAN &
  88 &
  \cellcolor[HTML]{D2B48C}21.46 &
  \multicolumn{2}{c}{1447.6} &
  \cellcolor[HTML]{D2B48C}188.2 &
  440 &
  \cellcolor[HTML]{D2B48C}57.2 &
  220 &
  \cellcolor[HTML]{D2B48C}28.6 \\\hline
\end{tabular}
\end{adjustbox}
\caption{Random and Longest incur negligible time and compute cost on our setups and hence, they are not included here. For all other strategies, the effective cost of data selection is non-trivial in comparison to training on the full-dataset. In three out of four strategies, it is possible to overshoot the cost of finetuning on the full dataset.
}
\label{tab:cost-estimates}
\end{table*}

A strong motivation for designing instruction selection strategies, and more broadly, data selection strategies draws from the need to train competitive models efficiently, both in terms of time and resource consumption. While the advantages towards this goal are more explicitly observed when source datasets are very large (pretraining datasets of the order of billions of tokens), instruction tuning datasets are typically much smaller in magnitude and thus the efficiency gains of selection can be less obvious to gauge. Accordingly, we evaluate if the proposed selection strategies \textit{consistently} provide this intended benefit by comparing the effective cost of selection against the performance of \selectedmodels{} and \fulldatamodels{}. 

\noindent \textbf{Setup} We compute the Cost of Selection as a product of the per-hour cost to user for renting a fixed compute infrastructure and the wall clock run time for running the selection for that strategy end-end. \S\ref{app:cost-breakdown} describes the full details of this computation including the wall clock time of running each selection (Table \ref{wall-clock-time}), while the total cost to user in summarized in Table \ref{tab:cost-estimates}. In Figure \ref{fig:cost-performance-correlation}, we plot the cost of selection per dataset compared to the performances of \selectedmodels{} on IFEVAL (all budgets are included in \S\ref{app:sec:cost-versus-perf} in the Appendix). 

\begin{figure*}[ht]    \includegraphics[width=\linewidth]{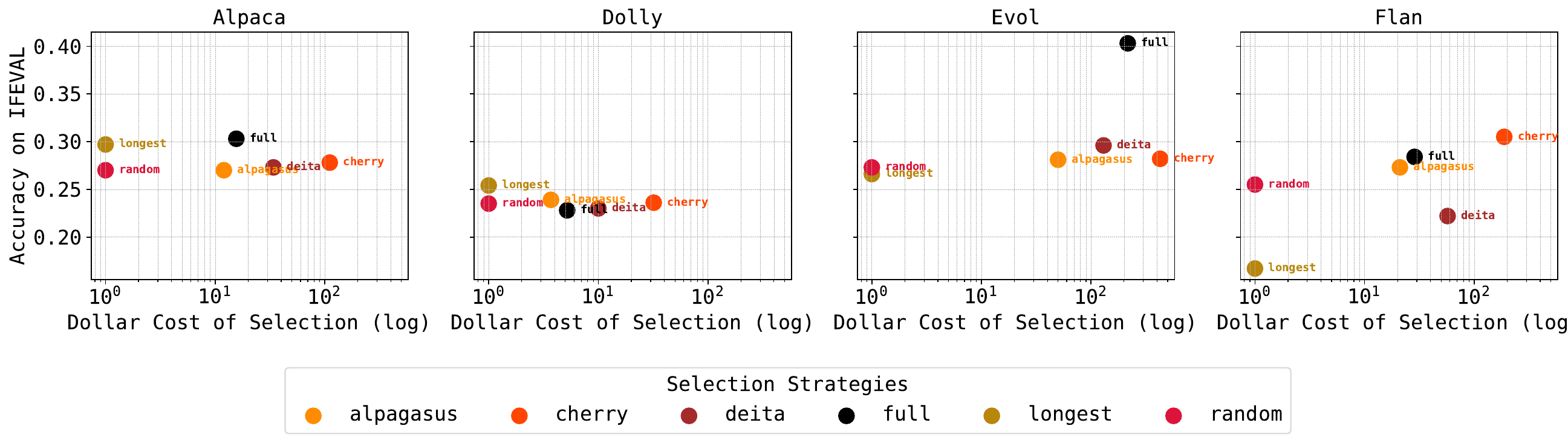}
    \caption{Cost Versus Performance Trade-Off at Selection Budget of 1000: Rather than reporting the average performance of random - we report the \textbf{lowest} performance amongst all our random trials to give the most pessimistic estimate of the performance of our random baseline.}
    \label{fig:cost-performance-correlation}
\end{figure*}
 % Note that since the entire dataset needs to be scored in accordance with a selection strategy's heuristic irrespective of the final sampling budget - the total cost of selection does not vary significantly with the change in the sampling budgets that we consider. 

\noindent\textbf{Finding} The effective cost of selecting data can often overshoot the cost of finetuning \fulldatamodels{} in some cases and the gains achieved through selection are marginal in comparison to the additional cost expended at carrying out the selection. While one potential cause of this could be the lack of more aggressive strategy-specific hyperparameter tuning, that is impractical for multiple reasons; For one, hyperparameter tuning in this space involves tuning for strategy dependent parameters such as the similarity threshold, $\lambda$ in \deita, the number of pre-experienced samples in \cherry, etc. in addition to traditional model training parameters like learning rate, scheduler and batch size.  Jointly optimizing for both these class of hyperparameters can significantly bloat the set of combinations to explore for hyperparameter optimization thus significantly increasing the cost of tuning. Secondly, under a practical setup where an NLP pracitioner expects to choose the best selection strategy \textit{amongst} several candidate strategies, a hyperparameter sweep for each candidate strategy would mandate tuning \textbf{all the strategies being examined}. From \ref{tab:cost-estimates}, this would imply summing the cost estimates across any row. We can clearly see that such an estimate would quickly overshoot the cost of full finetuning for any strategy. 

One interesting and consistent observation from this cost-benefit analysis is the surprising performance gain shown by \selectedmodels{} over \fulldatamodels{}. Both, \selectedmodels{} and \randombaselines{} often beat the \fulldatamodels{} across several experimental configurations. While some of these gains may be attributed to the lack of hyperparameter tuning for \fulldatamodels{}, supporting evidence from literature in the space of instruction data selection (\cite{qin2024unleashingpowerdatatsunami, Zhou2023LIMALI, Zhao2024LongIM, ge2024clusteringrankingdiversitypreservedinstruction} does imply that training on selected data can be beneficial (even though not necessarily cost effective). Empirically, this is also visible from the performance of our \strictrandom{} baselines: through the majority of our evaluation, the \strictrandom{} baselines underperform all other strategies indicating that systematically excluding datapoints that are selected by selection strategies definitely harms performance. 

\begin{takeaway}
\textit{\textbf{Takeaways}} Models can be trained to follow user instruction with relatively small subsets of data. It remains unclear though, if this selection is significantly more performant and cost-effective if carried out using selection strategies other than naive random sampling.  
\end{takeaway}

\section{Discussion and Conclusion}
This work demonstrates that selection strategies are not consistently competitive across setups and \textbf{this puts them at a risk of falling short of even random sampling} under a wider range of instruction tuning datasets, selection budgets and benchmarks. We also highlight that selection cost often surpasses the cost of full fine-tuning, without consistently delivering proportional benefits. 

\paragraph{Random Baselines offer consistency, reasonable and cost-effective performance:} Our conclusions on the performance of random baselines in this setting can be considered aligned to contemporary work demonstrating the unreasonable effectiveness of random baselines in several other domains; \citet{Yauney2024StrongerRB} discuss the significant competence of maximum expectancy random baselines in in-context learning by highlighting how standard random baselines may be severely underestimated on validation sets that are smaller in size. Similarly, \citet{Lu2023StringsFT} find that random baselines for prompt optimization can prove to be effective separators for prompt-style classification even challenging the assumptions that mandate task relevance and human readability in such tasks. Accordingly, our construction of random baselines must improve at scale to get a realistic calibration of the performance of our proposed methods.  

\paragraph{Instruction selection performance claims do not stand agnostic to the adopted experimental configurations} This dependence significantly harms their ease of adoption. Conversely, proposed instruction selection strategies may be more usable to NLP practioners if the efficacy of methods are tested across a wider range of experimental parameters (more budgets, datasets of differing distributions, etc.).

\paragraph{The Limits of \textit{Selective} Training in \textit{General}-Purpose Instruction Following}
General purpose instruction following is an unbounded recall problem as it can involve a fairly vast set of capabilities depending upon the context. There isn't a clear consensus on what are the sufficient conditions for claiming competence in general purpose instruction following: Models trained on selected data may show performance improvement against few limited facets but degrade it on unseen ones. Even using automatic metrics that act as proxies for human judgement seems unreliable as these metrics are also fallible \cite{Zheng2024CheatingAL} and susceptible to bias \cite{panickssery2024llmevaluatorsrecognizefavor}. Finally, since instruction following has evolving expectations, standardizing the choice of evaluation through human corroboration may only provide a stopgap solution \cite{vandermeer2024annotatorcentricactivelearningsubjective, Shen2023LargeLM}. As the complexity of such evaluation can be simplified for known test distributions, selection design effort may be more reliable and consistent in such fields.

% standardizing the choice of one of the existing instruction following benchmarksGeneral Purpose Instruction Following lacks a standard evaluation criterion; Consequently, selection strategcannot be

% through a human evaluation that corroborates a high correlation between human judgements would only provide a stopgap solution in our understanding due to the evolving and subjective expectations of 

% is intended to aid our calibration of how well instruction selection strategies generalize to realistically similar experimental configurations. Our experiments 

% We believe that the observed brittle generalization reduces the usability of such techniques for any NLP practioner who wishes to adopt data selection as a means to train competitive models in a resource efficient manner (as was the original intention of the conceptualization of these strategies \cite{qin2024unleashingpowerdatatsunami}). Our key takeaways are summarized below.

\section*{Limitations}
Since our work's goal is study the competency of models on a highly subjective goal, general purpose instruction following, conducting a comprehensive human evaluation to support our conclusions was not feasible. Our work also does not address other attributes that selection strategies differ by: including but not limited to the use of different base models, their impact on tuning strategies (like preference optimization) and alignment objectives. 

\section*{Ethics Statement}
This work highlights the potential of availing negative utility in the field of instruction data selection. Through the evidence in this work, we encourage a more conscious allocation of compute and dollar cost to reduce unnecessary computational overheads. Our code base and training logs (to validate wall clock times) will be released under the MIT License. 
% the quality of annotations are few of the many complexities associated with human annotation
% \hdtodo{Human Evaluation is missing though we don't think it is a scalable solution: In this space amortizing the subjective judgement criterion adopted by humans, covering representative annotator demographics and mitigating experimental biases that impact the quality of annotations are few of the many complexities associated with human annotation. These difficulties can require significant resource and technical investment to overcome and even once established, the evolving expectations of instruction following models could mean that such setups need to be updated with time, requiring recurring expense and supervision.}
\bibliography{custom}

\appendix

\section{Appendix}

\label{sec:appendix}
\subsection{Hyperparameter Configurations}
\label{hyperparameters}

We do our evaluations across 3 setups, trying to maximize the coverage of training setups that have been adopted by the strategies we reproduce. Additionally, we carried out one evaluation with LORA \cite{Hu2021LoRALA} to test if some weak correlation about the performance trends of selection strategies could be gleaned from low-rank finetuned models. The results for that evaluation are shown in  \S\ref{fig:lora-fft-correlation}. We present results from the hyperparameter configurations that matches the MMLU performance of reported for each strategy. The standard deviation with reported numbers along with confidence values for our hyperparameter runs across MMLU are provided in Table \ref{tab:confidence}. Since the work we study did not report \ifeval{}, \llmbar{} or \alpacaeval{} length-controlled win-rates - we were only able to utilize MMLU numbers (reported by all) as our sanity check for replication. \\ 

To replicate \cherry{}, we used the code open-sourced by the authors on \href{https://github.com/tianyi-lab/Cherry_LLM}{Github}, making minor adaptations to add support for new datasets. Following the default setup advised in \cite{Li2023FromQT}, we train our pre-experienced model for 1000 samples using the training configurations specified by the authors. For \deita{} also, we adopt the code opensourced by the authors on \href{https://github.com/hkust-nlp/deita}{Github}. We use the Mistral-7B-v0.1 for embedding generation, along with the \href{https://huggingface.co/hkust-nlp/deita-quality-scorer}{quality scorer}. For our similarity metrics, we used the same distance metric: cosine but different thresholds as keeping the default threshold led to an underflow for few of the models. We carry out inference using \href{https://github.com/vllm-project/vllm}{VLLM} to improve efficiency of our inference in \deita{}.
\begin{table*}[]
\centering
\begin{tabular}{cccccccc}
\hline
\textbf{Setup} &
  \textbf{LR} &
  \textbf{Optimizer} &
  \textbf{BS} &
  \textbf{MSL} &
  \textbf{Epochs} &
  \textbf{\begin{tabular}[c]{@{}c@{}}Warmup\\ Ratio\end{tabular}} &
  \textbf{\begin{tabular}[c]{@{}c@{}}LR \\ Scheduler\end{tabular}} \\ \hline
\textbf{Set 2}  & 2e-5 & Adam & 128 & 512  & 3 & 0.03 & Cosine \\ \hline
\textbf{Set 1}  & 2e-5 & Adam & 128 & 512  & 3 & 0.03 & Linear \\ \hline
\textbf{Set 3}  & 1e-5 & Adam & 128 & 512  & 3 & 0.03 & Linear \\ \hline
\textbf{Set 4} [LORA] & 2e-5 & Adam & 128 & 1024 & 5 & 0.3  & Linear \\ \hline
\end{tabular}
\caption{Hyperparameter Configurations for our experimental setup}
\end{table*}

\begin{table*}[]
\centering
\small
\begin{tabular}{llccccc}
\hline
\multicolumn{1}{c}{\textbf{Dataset}} &
  \multicolumn{1}{c}{\textbf{Strategy}} &
  \multicolumn{1}{c}{\textbf{\begin{tabular}[c]{@{}c@{}}MMLU\\ (Set 1)\end{tabular}}} &
  \multicolumn{1}{c}{\textbf{\begin{tabular}[c]{@{}c@{}}MMLU\\ (Set 2)\end{tabular}}} &
  \multicolumn{1}{c}{\textbf{\begin{tabular}[c]{@{}c@{}}MMLU\\ (Set 3)\end{tabular}}} &
  \multicolumn{1}{c}{\textbf{\begin{tabular}[c]{@{}c@{}}Standard \\ Deviation\end{tabular}}} &
  \multicolumn{1}{c}{\textbf{\begin{tabular}[c]{@{}c@{}}Confidence\\ Interval\end{tabular}}} \\ \hline
alpaca & alpacasus & 0.351 & 0.352 & 0.345 & 0.004 & 0.012 \\ \hline
evol   & cherry    & 0.354 & 0.354 & 0.348 & 0.003 & 0.011 \\ \hline
evol   & longest   & 0.352 & 0.352 & 0.349 & 0.002 & 0.005 \\ \hline
flan   & alpacasus & 0.351 & 0.351 & 0.347 & 0.002 & 0.007 \\ \hline
dolly  & alpacasus & 0.352 & 0.352 & 0.347 & 0.003 & 0.009 \\ \hline
alpaca & longest   & 0.353 & 0.351 & 0.345 & 0.004 & 0.013 \\ \hline
evol   & alpacasus & 0.348 & 0.348 & 0.349 & 0.001 & 0.002 \\ \hline
flan   & cherry    & 0.344 & 0.343 & 0.344 & 0.001 & 0.002 \\ \hline
dolly  & longest   & 0.348 & 0.347 & 0.345 & 0.002 & 0.005 \\ \hline
dolly  & cherry    & 0.348 & 0.349 & 0.345 & 0.002 & 0.006 \\ \hline
alpaca & cherry    & 0.346 & 0.345 & 0.344 & 0.001 & 0.003 \\ \hline
flan   & longest   & 0.345 & 0.345 & 0.345 & 0.000 & 0.000 \\ \hline
\end{tabular}
\caption{MMLU Values for Budget 1000 across all hyperparameter setups. Since we saw the highest (relative) correlation between all benchmarks at this budget, we chose the final hyperparameter set based on this budget's value.}
\label{tab:confidence}
\end{table*}

\begin{table*}[h]
\centering
\begin{tabular}{lcc}
\hline
\textbf{Paper} & \textbf{\begin{tabular}[c]{@{}l@{}}Reported \\ Value\end{tabular}} & \textbf{\begin{tabular}[c]{@{}l@{}}Our Value\\ (Budget - 1k)\end{tabular}} \\ \hline
\textbf{Alpagasus  at 9K} & 36.93 & 35.2 \\ \hline
\textbf{Cherry at 3.5K}   & 36.51 & 35.2 \\ \hline
\textbf{Cherry at 7K}     & 33.08 & 35.2 \\ \hline
\end{tabular}
\caption{Reported Performance versus replicated performance; While \deita also used the same base model as us, they use a 13B parameter model and hence, we do not compare with their numbers. Set 2 was the closest in evaluation to these numbers so we chose Set 2 for reporting our results.}
\label{reported}
\end{table*}
% \section{Strategy dependent Hyperparameter Tuning}
% \begin{table}[]
% \begin{tabular}{lrl}
% \textbf{Dataset} & \multicolumn{1}{l}{\textbf{Similarity Threshold}} & \textbf{Wall Clock Time} \\
% \textbf{Alpaca} & 0.96 & $\sim$30 minutes \\
% \textbf{EVOL}   & 0.98 & 10 minutes       \\
% \textbf{FLAN}   & 0.98 & 15 minutes       \\
% \textbf{DOLLY}  & 0.98 & $\sim$8 minutes 
% \end{tabular}
% \end{table}

\section{Detailed Cost Estimation Across Data Selection Budgets}
\label{app:cost-breakdown}
All our estimates are provided assuming the following infrastructure: 8 A6000s, 128 CPUs provided by \href{https://cloud.google.com/compute/all-pricing}{Google Cloud Estimator}. The Dollar Cost of renting our infrastructure per hour is about 8 USD. 
% \begin{figure}[h]
%     \centering
%     \includegraphics[width=0.9\linewidth]{assets/cloud-provision-cost-estimate.png}
%     \caption{Cost Estimate for Renting Infrastructure on Google Cloud. For our setup, we calculate the rent cost at \~8 USD per hour.}
%     \label{fig:google-cloud-estimate}
% \end{figure}

 A detailed breakdown of the costs associated with each step of the selection is provided in Table \ref{wall-clock-time}. Note that the cost of selection doesn't vary significantly with the change in the selection budget as the entire dataset needs to be sorted in accordance with the strategy guided metric, irrespective of the final budget. 
\begin{table*}[]
\centering

\begin{tabular}{cc}
\rowcolor[HTML]{CBCEFB} 
\textbf{\Sstrategy} & \textbf{Wall Clock Time on Rented Infrastructure (hr)}  \\ \hline
\alpacasus &  0  \\ \hline 
\longest & \textcolor{purple}{Total Time per 1000 samples: 1 minute} \\ \hline 
\cherry &
  \begin{tabular}[c]{@{}c@{}}1.457 minutes minutes for 1000 samples embedding construction \\ + 7 mins for training pre-experienced model on 1000 samples (One-time cost, so ignored) \\ +  15 minutes for computing token loss over 1000 samples \\
\textcolor{purple}{Total Time per 1000 samples: 16.45 minutes} \end{tabular} \\ \hline 
\deita  & \begin{tabular}[c]{@{}c@{}}2 minutes for 1000 samples for embedding construction (Mistral 7B) +\\120 minutes for scoring 1000 samples w/o VLLM  \\2 minutes for scoring 1000 samples with VLLM +\\ 1\footnote{This time varies considerably depending upon the dataset used and we use the shortest time it takes for an optimistic estimate} minute \textit{at least} for filtering 1000 sample. \\ \textcolor{purple}{Total Time per 1000 samples: 5 minutes} \end{tabular} \\\hline 
\end{tabular}
\caption{Wall Clock Times for Each Selection Strategy: We offset the time of computatation we subsample 100K samples from EVOL and then select samples from that subset.}
\label{wall-clock-time}
\end{table*}

\subsection{Estimating Performance Using Cost-Effective Proxies}

\begin{figure}[h]
    \small
    \centering \includegraphics[width=1\linewidth]{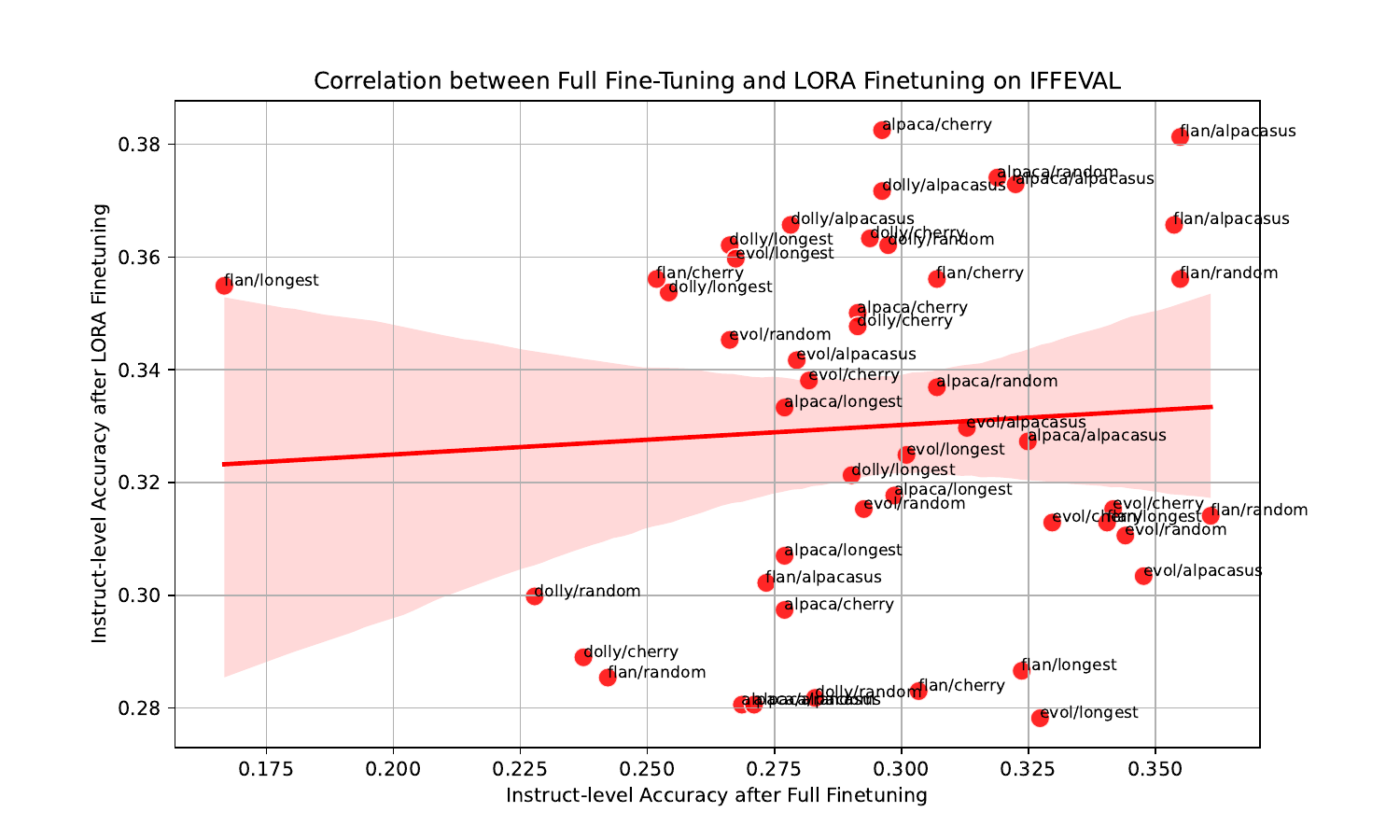}
    \caption{There isn't any observable correlation between the performance of models finetuned with and without LORA across our setups indicating that we cannot reliably predict the optimal selection strategy on a faster, cost-effective parameter efficient setup.}
    \label{fig:lora-fft-correlation}
\end{figure}

While it is not possible to largely modify the cost of a selection strategy, it might be possible to offset the cost of finetuning the models on subsets generated via different selection strategies through parameter efficient techniques. If such trends are correlated with the performance of the selected data on the full variant of the model, NLP practioners can potentially design a set of relatively low-cost experiments to rapidly identify the optimal selection strategy to further carry out their selection. Recent work like \cite{Xia2024LESSSI}, even leverage such correlation to achieve great efficiency in task-specific instruction selection. For preliminary experimentation, we rerun all our experiments with the modification of including LORA modules in our finetuning. This reduces the memory footprint of training by to only 0.0038 times of the memory footprint of full finetuning along with faster training by half of its full-finetuning counterpart. In \ref{fig:lora-fft-correlation} we plot the correlation between the instruct-level-accuracy on IFEVAL for models trained with and without LORA. However, we don't find any reasonable correlation between these performances highlighlting a need to identify cost-effective methods of predicting the suitability of a custom budget and source distribution to a given selection strategy.

\subsection{Benchmark Evaluations for All Configurations}
\label{app:alpacaeval}
\begin{figure}
    \centering
    \includegraphics[width=\linewidth]{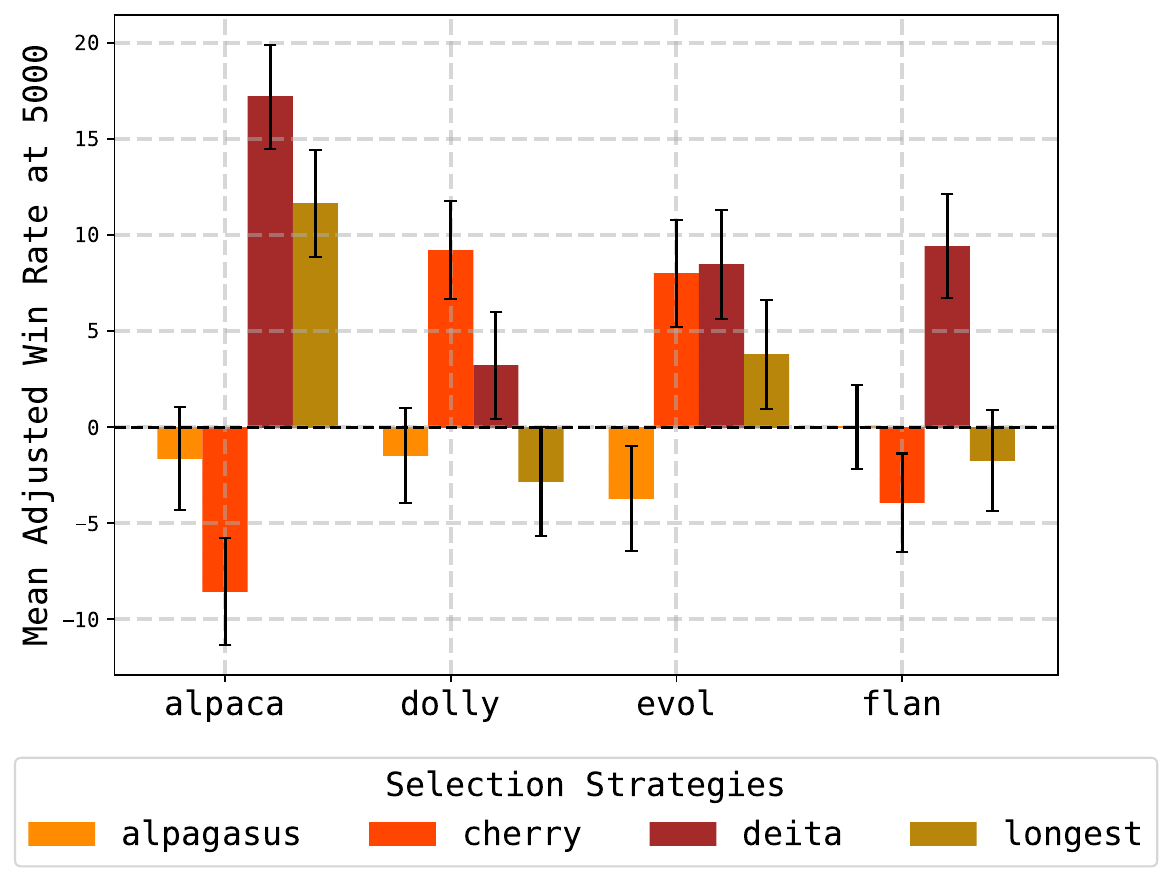}
    \caption{AlpacaEval Length Controlled Win Rate at 5K: Models do not show a consistent trend in performance when scaled from 1000 samples or across datasets. }
    \label{fig:alpacaeval-5k}
\end{figure}

\label{sec:openllm}
\begin{figure*}[h]
    \centering
    \begin{subfigure}[b]{\linewidth}
        \includegraphics[width=\linewidth]{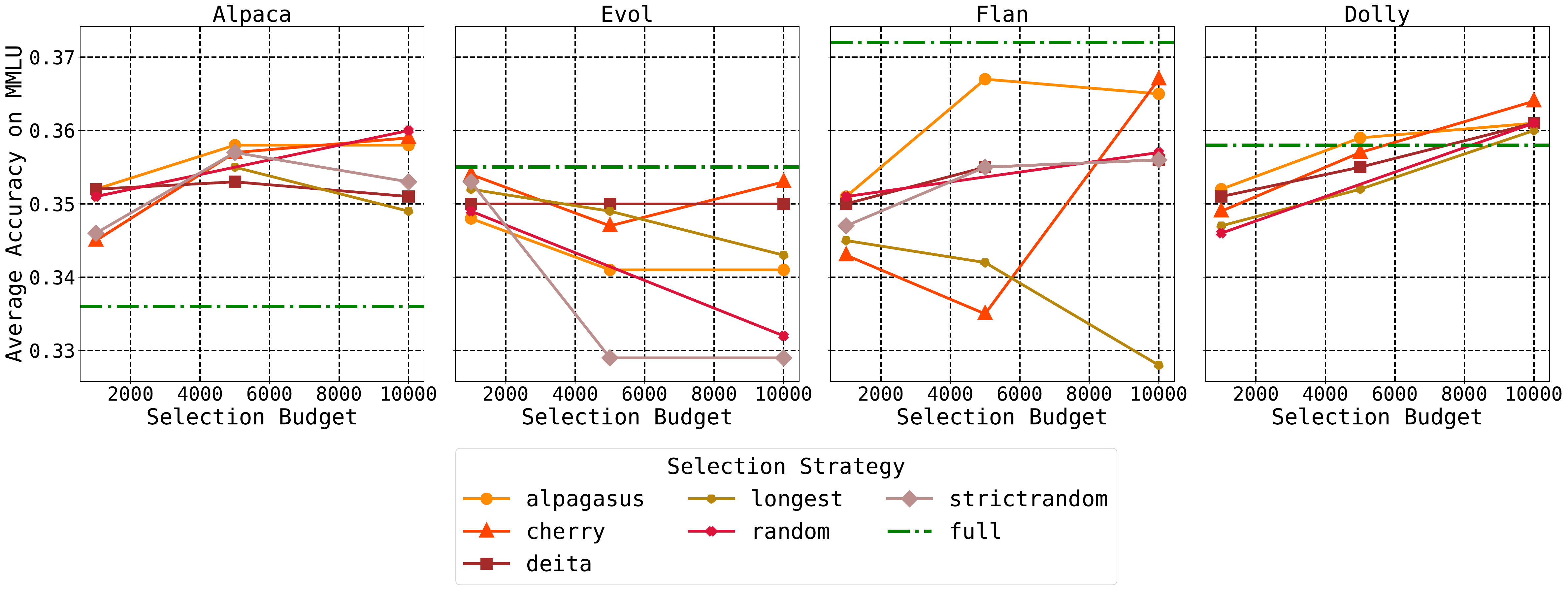}
        \caption{\randombaselines versus \selectedmodels on MMLU}
        \label{fig:mmlu-all}
    \end{subfigure}
    \hfill 
    \begin{subfigure}[b]{\linewidth}
        \includegraphics[width=\linewidth]{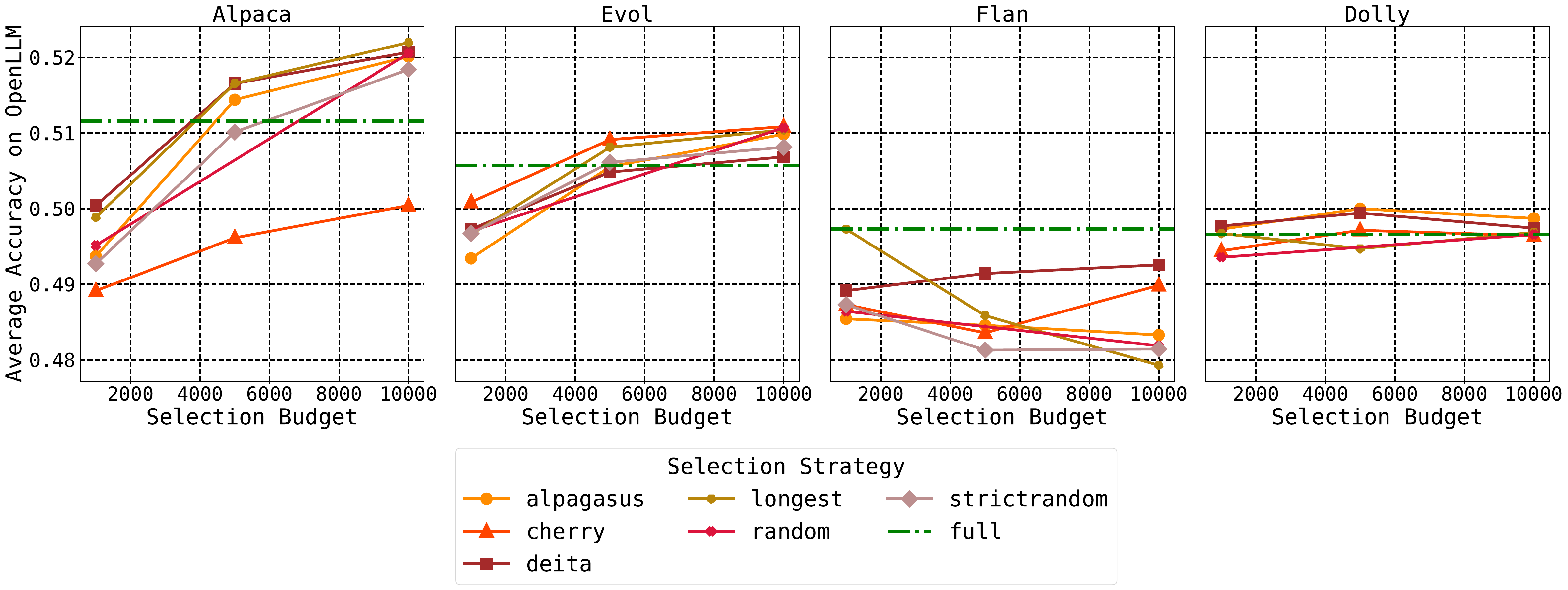}
        \caption{\randombaselines versus \selectedmodels on OPENLLM}
        \label{fig:openllm-all}
    \end{subfigure}
    \caption{Comparison of \randombaselines and \selectedmodels on different models}
    \label{fig:combinedfigs}
\end{figure*}

\label{sec:iffeval}
\begin{figure*}
    \centering
    \includegraphics[width=\linewidth]{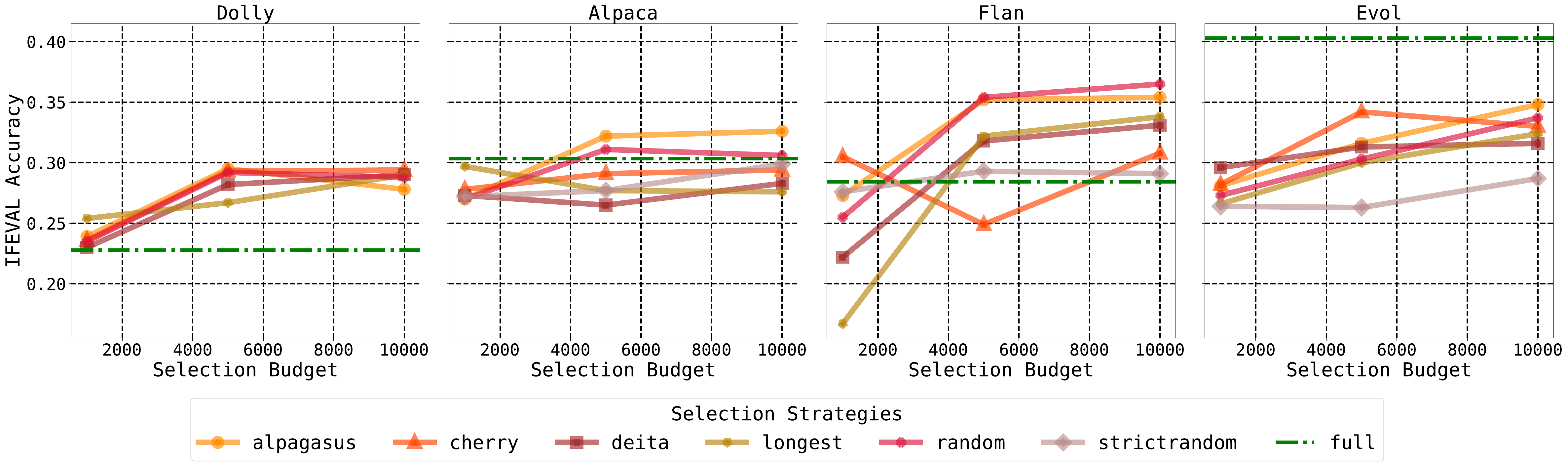}
    \caption{Performance Comparison between \randombaselines and \selectedmodels on IFEVAL: We report the average of all random runs (both random and strict random) for a particular configuration in any result. }
    \label{fig:iffeval-all}
\end{figure*}

% \subsection{AlpacaEval Full Results}

% \begin{figure}
%     \centering
%     \includegraphics[width=0.5\linewidth]{assets/alpacaeval_winrate_random_versus_full.pdf}
%     \caption{AlpacaEval versus Full }
%     \label{app:alpacaevalvsfull}
% \end{figure}

\subsection{Correlation between all Benchmarks}
\label{app:sec:cost-versus-perf}

\begin{figure*}
    \centering
    \includegraphics[width=1\textwidth]{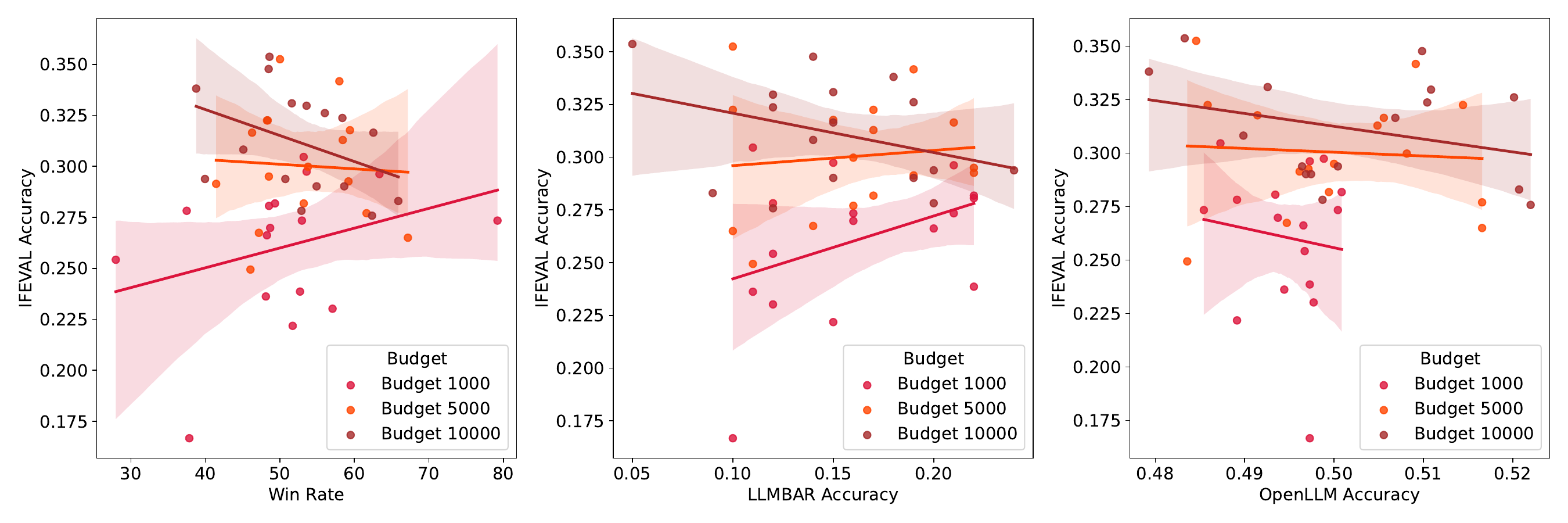}
    \caption{Correlation between Benchmarks: All benchmarks show poor correlation, especially at larger budgets (almost negatively correlated performances)}
    \label{fig:correlation-btw-benchmarks}
\end{figure*}

\subsection{Cost Versus Performance Trade-Offs for All Budgets}
\begin{figure*}
    \centering
    \includegraphics[width=\textwidth]{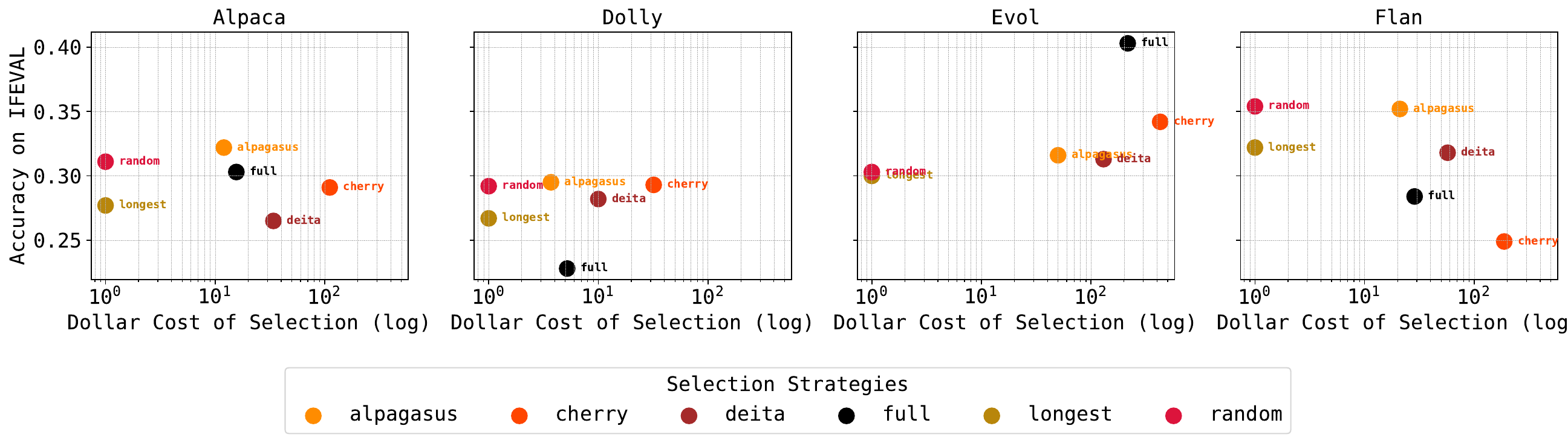}
    \caption{Cost Versus Performance Comparison at 5K budget: We highlight that random performs competitively across most setups }
    \label{fig:cost-vs-perf-5k}
\end{figure*}

\begin{figure*}
    \centering
    \includegraphics[width=\textwidth]{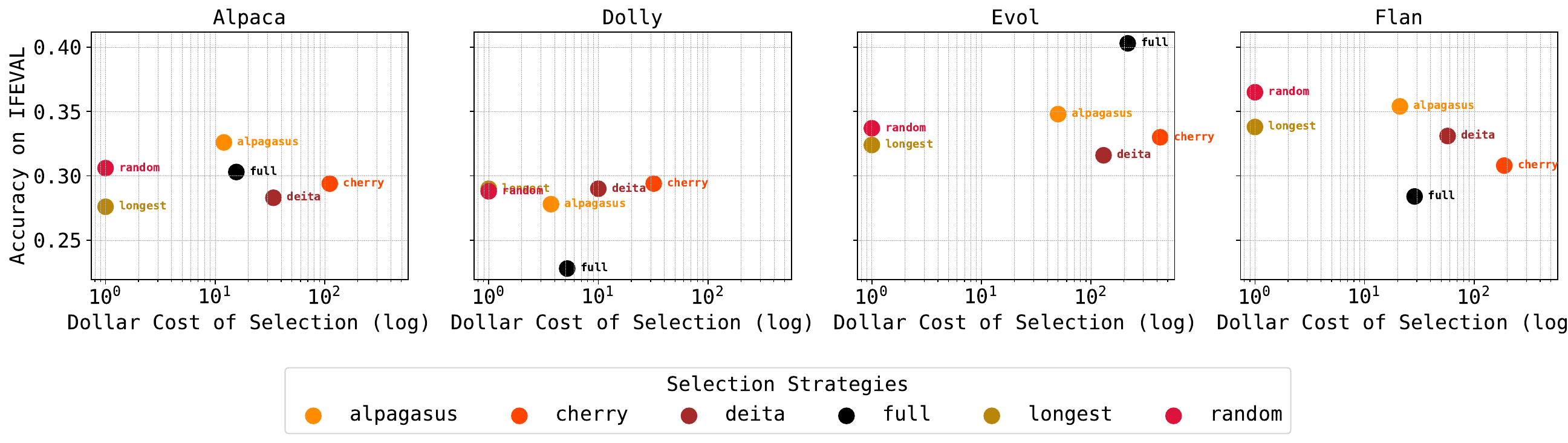}
    \caption{Cost Versus Performance Comparison at 10K:We highlight that random performs competitively across most setups }
    \label{fig:cost-vs-perf-10k}
\end{figure*}

\end{document}

In this work, we establish that general purpose instruction selection strategies generalize poorly, often being unable to beat even random baselines consistently at even reasonably similar experimental setups.